\documentclass[
]{ceurart}
\usepackage[utf8]{inputenc}
\usepackage{listings}
\usepackage{makecell}
\usepackage{tabularx} 

\begin{document}

\copyrightyear{2025}
\copyrightclause{Copyright for this paper by its authors.
  Use permitted under Creative Commons License Attribution 4.0
  International (CC BY 4.0).}

\conference{CLEF 2025 Working Notes, 9 -- 12 September 2025, Madrid, Spain}

\title{StylOch at PAN: Gradient-Boosted Trees with Frequency-Based Stylometric Features}
\title[mode=sub]{Notebook for the PAN Lab at CLEF 2025}


\author[1,2,3,4]{Jeremi K. Ochab}[
orcid=0000-0002-7281-1852,
email=jeremi.ochab@uj.edu.pl,
]
\cormark[1]

\author[3]{Mateusz Matias}[
email=mateusz.matias@student.uj.edu.pl,
]

\author[3]{Tymoteusz Boba}[
email=tymoteusz.boba@student.uj.edu.pl,
]

\author[4]{Tomasz Walkowiak}[%
orcid=0000-0002-7749-4251,
email=tomasz.walkowiak@pwr.edu.pl,
]

\address[1]{Institute of Theoretical Physics, Jagiellonian University, Krak\'{o}w, Poland}
\address[2]{M. Kac Center for Complex Systems Research, Jagiellonian University, Krak\'{o}w, Poland}
\address[3]{Faculty of Physics, Astronomy and Applied Computer Science, Jagiellonian University, Krak\'{o}w, Poland}
\address[4]{Faculty of Information and Communication Technology, Wroclaw University of Science and Technology, Wroclaw, Poland}

%


\cortext[1]{Corresponding author.}

\begin{abstract}
 This submission to the binary AI detection task is based on a modular stylometric pipeline, where: public spaCy models are used for text preprocessing (including tokenisation, named entity recognition, dependency parsing, part-of-speech tagging, and morphology annotation) and extracting several thousand features (frequencies of n-grams of the above linguistic annotations); light-gradient boosting machines are used as the classifier.
We collect a large corpus of more than 500 000 machine-generated texts for the classifier's training.
We explore several parameter options to increase the classifier's capacity and take advantage of that training set.
Our approach follows the non-neural, computationally inexpensive but explainable approach found effective previously.

\end{abstract}


\maketitle

\section{Introduction}

The rapidly developing landscape of Large Language Models (LLMs) has revolutionised natural language processing (NLP), enabling the use of machine-generated texts (MGTs) throughout society on a daily basis. 
The use of these tools, especially in some professional environments such as academic \citep{lund2023chatgpt} , medical \cite{de2023chatgpt}, legal, or news reporting, raises concerns around issues of plagiarism, factual reliability, and many others. 
The \textit{``Voight-Kampff'' Generative AI Authorship Verification Task}~\cite{bevendorff2025b} at the PAN and ELOQUENT 2025 workshop~\cite{bevendorff2025a}, and specifically Subtask 1 ``AI Detection Sensitivity'' answers the urgent need to develop reliable model detection.
The subtask is a classical binary text classification task, i.e. categorising a given text as a human or machine written. The additional challenge comes from changing the style of MGTs, mimicking specific human authors, testing on unseen models, and using obfuscation strategies.
In submission to this subtask, we strive to expand on the simplistic non-neural feature-based classifiers that were previously found effective, using boosted trees with stylometric (linguistically explainable) features.

\section{Background}
\subsection{MGT detection methods}
There is a considerable variety of MGT detection methods reviewed in \cite{crothers_machine-generated_2023,wu_survey_2025}, but also in the overview of last year's \textit{Voight-Kampff Generative AI Authorship Verification Task} at PAN and ELOQUENT 2024 \cite{bevendorff_overview_2024}.
They included systems based on (i) terms, (ii) perplexity or logit statistics, (iii) watermarking, and their mixtures.
The watermarking approach relies on embedding an imperceptible signature in the generated texts at some stage of the generator training, text generation, or post-processing that modifies character- or word-level distributions. In the present submission, we disregard this approach due to the task's constraints. 
The logits statistics approach typically involves zero-shot white-box methods, i.e., ones that require access to the LLM generator (or its surrogate) in order to compute either the likelihood of a text being generated by it or features later used in a classifier.
The black-box alternatives, instead,  would machine-regenerate a given text sample and subsequently compare it to the original to obtain a similarity score.
Finally, term-based systems are typically neural fine-tuned classifiers (from the BERT family with modifications) using word embeddings or linguistic (stylometric) features. 

Our submission follows works such as \cite{guo_machine-generated_2024,miralles_team_2024,yadagiri_team_2024,guo_blgav_2024}, which either utilised various stylometric features, augmented data, or expanded the training dataset.
We especially find our approach similar to the simple SVM classifier on the TF-IDF features \cite{lorenz_baselineavengers_2024}, which outranked all neural baselines and most neural-based submissions in the last year's task.
Classifiers based on stylometric features were also found to be effective elsewhere \cite{opara_styloai_2024}.


\subsection{MGT detection robustness}

The performance of MGT detection can generally degrade due to two factors: out-of-distribution issues and attacks~\cite{wu_survey_2025}.
The former encompasses generalisation issues such as: cross-domain (involving changing text type, and consequently its vocabulary, style, topics, etc.), cross-language (involving not only switching the language of the text but also linguistic interference due to non-nativity of the authors) and cross-LLM (involving detection of text generators unseen during the detector's training).
The latter includes: paraphrasing output of one LLM by another (therefore changing the textual feature distribution of the former)~\cite{sadasivan_can_2025}, adversarial text perturbations on different levels (characters \cite{stiff_detecting_2022}, syntax, \cite{bhat_how_2020} or lexis, \cite{crothers_adversarial_2022}), prompt engineering (taking advantage of in-context learning to change LLM's characteristics by varying prompts \cite{guo_how_2023,liu_detectability_2024} including mimicking specific authors or character profiling~\cite{przystalski_building_2025}) and other attacks.

Reportedly~\cite{sarvazyan_supervised_2023}, supervised detectors can generalise reasonably well across LLM scales but less so across model families.
On the other hand, issues that were found to be challenging were incorporating unseen languages and performing simple attacks such as Unicode obfuscation or shortening text length ~\cite{bevendorff_overview_2024}.
Our own approach has been found vulnerable to cross-domain detection \cite{przystalski2025stylometry} as tested on \cite{sarvazyan_overview_2023}, but on a closed domain it was robust to one-step paraphrasing.
Furthermore, the unexpected performance of the aforementioned SVM TF-IDF classifier \cite{lorenz_baselineavengers_2024} was mainly due to its robustness to obfuscation.

We did not explicitly design our detector to target any of these issues; however, we follow the general recommendation~\cite{wu_survey_2025} that supervised detectors can effectively defend against some of them by continually expanding training datasets (with adversarial examples, examples of LLM families, examples of text types, etc.) and fine-tuning even on small samples.

\section{System Overview}

In general, our submission employed (i) gradient-boosted tree models together with (ii) feature engineering and, crucially, (iii) a large training dataset.
We did not target any obfuscation techniques.
Similarly to our previous work \cite{przystalski2025stylometry,argasinski2024}, we used a modular Python pipeline for interpretable stylometric analysis being developed for CLARIN-PL\footnote{\url{https://gitlab.clarin-pl.eu/stylometry/cl_explainable_stylo}}\cite{ochab2024}.
It is designed to connect text preprocessing and linguistic feature extraction with various existing NLP tools, classifiers, explainability modules, and visualisation.

\subsection{Data source}

Following our unremarkable attempt~\cite{przystalski2025stylometry} (F1 = 0.54 compared to an ensemble of stylometric features and transformers \cite{mikros2023transformers} and the highest ranked result 0.81) in cross-domain MGT detection on \textit{AuTexTification}~\cite{sarvazyan_overview_2023} benchmark -- where training was performed on tweets, how-to articles and legal documents, while testing on reviews and news -- we decided that our model needs as comprehensive and varied training data as possible in order for the validation result to hold on test set.

For that purpose, we have collected in total 563 571 text samples from several openly accessible datasets~\cite{bevendorff2025b,sarvazyan_overview_2023,yu_cheat_2025,guo_how_2023,su_hc3_2024,li_mage_2024,macko_multitude_2023,wang_m4_2024} designed as benchmarks in MGT detection, see Table \ref{tab:data}.
The number reported above already takes into account dropouts due to issues with special characters or incompatibility of data structure that we were not able to solve within the time constraints of the PAN's task.
In particular cases, not all data were incorporated (e.g. training but not validation set in the case of \textit{AuTexTification} and \textit{PAN's Voight-Kampff Generative AI Detection}; consequently, not all available genres were included). 
Some of these datasets themselves were collected from other openly available datasets and augmented with the generated texts.
The total number of LLM labels available in that dataset was 348.

The source, genre and model labels were not used in the training.

\begin{table}[h]
\caption{Overview of datasets used in training. Items in parentheses refer to all the collected samples, while items without parentheses refer to the samples used in training.}
\label{tab:data}
\renewcommand\cellalign{tl}
\renewcommand\theadalign{tl}

\begin{tabularx}{\textwidth}{@{}>{\raggedright\arraybackslash}p{3cm} c c >{\raggedright\arraybackslash}p{1.5cm} >{\raggedright\arraybackslash}X@{}}
\toprule
\textbf{Dataset} & \textbf{Samples} & \textbf{Word Count} & \textbf{Genres} & \textbf{Models} \\
\midrule

\makecell[l]{PAN’25 Generative \\ AI Detection} & 
\makecell[c]{23 704 \\ (23 707)} & 
14 727 408 & 
\makecell[l]{essays \\ fiction \\ news} & 
\makecell[l]{\textit{human}, deepseek-r1-distill-qwen-32b \\ falcon3-10b-instruct, gemini-1.5-pro \\ gemini-2.0-flash, gemini-pro\\ gemini-pro-paraphrase, gpt-3.5-turbo\\ gpt-4-turbo, gpt-4-turbo-paraphrase\\ gpt-4.5-preview, gpt-4o, gpt-4o-mini \\ llama-2-70b-chat, llama-2-7b-chat \\ llama-3.1-8b-instruct, llama-3.3-70b-instruct\\ mistral-8b-instruct-2410\\ mistral-7b-instruct-v0.2\\ mixtral-8x7b-instruct-v0.1, o3-mini\\ qwen1.5-72b-chat-8bit, text-bison-002} \\

Autextification~\cite{sarvazyan_overview_2023} & \makecell[c]{21 832 \\ (21 832)} & 1 367 323 & \makecell[l]{news \\ reviews \\ (tweets \\ how-to \\ legal)} & 
\makecell[l]{\textit{human}, BLOOM-1B7, BLOOM-3B,\\ BLOOM-7B1, babbage, curie,\\ text-davinci-003} \\

CHEAT~\cite{yu_cheat_2025} & \makecell[c]{15 394 \\ (15 395)} & 165 584 & abstracts & gpt-3.5-turbo \\

HC3~\cite{guo_how_2023} & \makecell[c]{0 \\ (48 644)} & 12 492 921 & \makecell[l]{Q\&A\\finance\\ medicine\\Wikipedia} & \makecell[l]{\textit{human}, ChatGPT} \\

HC3 Plus~\cite{su_hc3_2024} & \makecell[c]{148 237 \\ (148 402)} & 11 250 436 & \makecell[l]{news \\ summaries \\ translations \\ question\\ paraphrases}  & \makecell[l]{\textit{human}, GPT-3.5-Turbo-0301} \\

MAGE~\cite{li_mage_2024} & \makecell[c]{318 958 \\ (319 071)} & 67 471 388 & \makecell[l]{opinions \\ reviews \\ news \\ Q\&A \\ stories \\ reasoning \\ Wikipedia \\ abstracts} & 
\makecell[l]{\textit{human}, gpt-3.5-turbo, text-davinci-002,\\ text-davinci-003, gpt\_j, gpt\_neox, opt,\\ flan\_t5, t0, bloom\_7b, GLM130B} \\

Multitude~\cite{macko_multitude_2023} & \makecell[c]{29 459 \\ (29 460)} & 6 175 907 &  \makecell[l]{news} & 
\makecell[l]{\textit{human}, alpaca-lora-30b, gpt-3.5-turbo\\ gpt-4, text-davinci-003, vicuna-13b\\
llama-65b, opt-66b, opt-iml-max-1.3b} \\

M4~\cite{wang_m4_2024} & \makecell[c]{5987} & XXX & \makecell[l]{Wikipedia \\ abstracts \\ peer reviews \\ news briefs} & 
\makecell[l]{\textit{human}, GPT-4, ChatGPT\\ text-davinci-003, Cohere\\ Dolly-v2, BLOOMz 176B} \\

\bottomrule
\end{tabularx}
\end{table} 

\subsection{Stylometric Features}

We considered two options: either a closed set of predefined but more interpretable features or an open set features generated programmatically but still partly based on linguistic analysis.

Regarding the first, when analysing our own Wikipedia-based dataset~\cite{przystalski2025stylometry}, we used StyloMetrix~\cite{okulska2023stylometrix}.
This open-source stylometric text analysis library calculates the appearance of 195 predefined features that include
grammatical forms (tenses, modal verbs, etc.), parts of speech, lexical items (types of pronouns, hurtful words, etc.), aspects connected to social media (e.g. sentiment analysis), syntactic forms, and general text statistics (e.g. type-token ratio).
StyloMetrix uses the spaCy model for English to extract these features. 
The classifiers based on this small feature set consistently scored lower than the alternative,
so the final submission comprised only the second option.

The second option follows the basic ideas used in the R package \texttt{stylo}~\cite{eder2016RJStylometry}, which is mainly computing token n-grams, but augmented with the various annotations.
At present, for preprocessing steps and said annotations (tokenisation, named entity recognition, dependency parsing, part-of-speech tagging, and morphology annotation) we use spaCy~\cite{ines_montani_2023_10009823} model \texttt{en\_core\_web\_lg}.
Specifically, we computed the normalised frequencies of:
\begin{itemize}
    \item lemmas (from uni- to trigrams), excluding named entities,
    \item part-of-speech tags (from uni- to quadrigrams) including punctuation,
    \item dependency-based bigrams (where token neighbourhood is defined by the distance in the dependency tree), excluding named entities,
    \item morphological annotations (unigrams) including entity types (i.e. using Named Entity Recognition to find named entities and replacing them with their types)
\end{itemize}
Each of the four feature classes could contain a maximum of 1500 items.
This particular set of features admittedly comes from some unresolved technical issues, but also from repeated trial and error on yet other authorial attribution datasets.
For instance, elsewhere~\cite{przystalski2025stylometry} we have found that punctuation features, such as the `SPACE' token, can detect human mistakes or artefacts in LLM processing or further data post-processing (a redundant whitespace character, e.g., at the beginning of a paragraph or a second one between words).
In that choice of feature classes we also try to minimise, although not strictly enforce, the generation of duplicate versions of the same feature in separate classes.

As presented in Table~\ref{tab:models} we also testes so-called \textit{culling} (i.e., ignoring features with \textit{document frequency} strictly higher or lower than the given threshold).
In the present submission, a majority of our models did not use culling.
In one case, we set the minimum document frequency to 0.1 (that is, about 50k out of 500k documents), which reduced the number of features from the initial 4594 to 3264.

\subsection{Classifier}

We take advantage of the existing solutions: Light Gradient-Boosting Machine (LGBM)~\cite{ke2017lightgbm} as the state-of-the-art boosted trees classifier and Scikit-learn~\cite{scikit-learn} for feature counting and cross-validation.
Following our previous experience on other, smaller datasets -- mainly in English and Polish languages -- during pipeline development, the LGBM classifiers parameters were set to:
\texttt{DART} boosting, \texttt{learning\_rate} = 0.5, enabled bagging (randomly selecting \texttt{bagging\_fraction} = 0.8 of data without resampling every \texttt{bagging\_freq} = 3 iterations).
At this time, we used the binary classifier, but it is possible -- and in fact in can be beneficial~\cite{przystalski2025stylometry} -- to train a multiclass model using the LLM labels, see Table~\ref{tab:data}, and then map it back to the binary `human vs. machine' labels.

The following parameters were used to produce separate submissions with increasing model capacity:
\begin{itemize}
    \item maximal number of leaves per tree (\texttt{num\_leaves}), 
    \item number of boosting iterations (\texttt{num\_iterations})
    \item maximal depth of the tree model (\texttt{max\_depth}), 
\end{itemize}
In our smaller pre-submission experiments (e.g., human authorship attribution on 2-100 novels, resulting in the number of samples of the order of thousands or tens of thousands at most) satisfactory results were obtained with \texttt{num\_leaves} = 5, \texttt{num\_iterations} = 100, \texttt{max\_depth} = 5.
We decided that with 500k text samples, 4k features and 348 LLM labels, the LGBM classifier required a higher capacity, hence we submitted three classifier versions: \textit{small}, \textit{medium} and \textit{big}, listed in Table~\ref{tab:models}.
Further hyperparameter optimisation is possible, but was not performed in the present submission.

Since LGBM training is fast, we used the stratified 10-fold cross-validation (CV) scheme to obtain more reliable validation and test error estimation.
We then decided to validate both a classifier from a single fold (\textit{-single}) and the probability scores averaged over classifiers trained on all CV folds (\textit{-cv}).

\begin{table}[]
\caption{Overview of submitted model parameters and their sizes. The \textit{Model size} refers to the size of model saved in a \texttt{.txt} file.}
\label{tab:models}
\begin{tabular}{cccccc}
\hline
\textbf{Model name} & \multirow{2}{*}{\textbf{Feature culling}} & \multicolumn{3}{c}{\textbf{LGBM parameters}} & \multirow{2}{*}{\textbf{Model size {[}kB{]}}} \\
  &   & \texttt{num\_leaves} & \texttt{num\_iterations} & \texttt{max\_depth} &   \\ \hline
\textit{small} & 0   & 10 & 100 & 8 & 229   \\
\textit{medium} & 0   & 12 & 500 & 10 & 851  \\
\textit{big} & 0   & 20 & 1500 & 12 & 3685   \\
\textit{culled} & 0.1   & 20 & 1500 & 12 & 3648   \\ \hline
\end{tabular}
\end{table}


\section{Results}
\subsection{Evaluation setup}

The environment for running and evaluating submissions to Subtask 1 ``AI Detection Sensitivity'' of the \textit{PAN: Voight-Kampff Generative AI Detection 2025} task was TIRA~\cite{froebe2023continuous}.
This platform allows dockerised submissions in order to ensure their reproducibility.
Upon submission our contribution was validated on two datasets, to which we refer as: "Validation 1" – the validation split of the dataset available for training (available texts and labels), "Validation 2" – the dataset used for evaluation at TIRA (not available to see its contents).
Both datasets could be used for classifier evaluation and selection; see Table~\ref{tab:res}.
TIRA platform produced the following six evaluation metrics (all on scale 0-1, with 1 representing the perfect score):
\begin{itemize}
    \item \textbf{ROC-AUC}: The area under the ROC (Receiver Operating Characteristic) curve
    \item \textbf{Brier}: The complement of the Brier score (equivalent to mean squared loss)
    \item \textbf{C@1}: A modified accuracy score that breaks ties by assigning \textit{non-answers} (class probability = 0.5) the average accuracy of the remaining cases
    \item $\mathbf{F_1}$: The harmonic mean of precision and recall
    \item $\mathbf{F_{0.5u}}$ : A modified $F_{0.5}$ measure (where precision weighs more than recall) that treats non-answers as false negatives
    \item The arithmetic mean of all above.
\end{itemize}
The final evaluation was also appended with the False Positive Rate (FNR) and False Negative Rate (FNR).
The submissions were ranked by a macro-average of the arithmetic mean over all individual data sources (all individual datasets contained in the test and the ELOQUENT collections). 


\subsection{Evaluation results}
Table~\ref{tab:res} presents the arithmetic mean scores of our submissions on the validation sets (available during submission) and the unobfuscated test set together with the best-ranked baseline and participant contribution \cite{macko2025mdokkinitrobustlyfinetuned}.
Both model capacity (model size and no feature culling) and cross-validation visibly led to higher scores on both datasets. 
The results from the obfuscated ELOQUENT dataset available at TIRA showed the same pattern in ROC-AUC metric, but there was no generally discernible dependence on the model size in the other metrics.
The detailed test results for the selected \textit{big-cv} model are shown in Table \ref{tab:final} (a).

\begin{table}[]
\caption{Mean performance on validation sets and the unobfuscated test set against the best baseline (TF-IDF) and the best contribution \cite{macko2025mdokkinitrobustlyfinetuned}. The values are arithmetic means of evaluation metrics.}
\label{tab:res}
\begin{tabular}{lccc} 
\toprule
\textbf{Approach} & \textbf{Validation 1} 
& \textbf{Validation 2} & \textbf{Test} \\
\midrule
\textit{small-single} & 0.943 & 0.885  
& 0.885 \\
\textit{medium-single} & 0.967 & 0.93 
& 0.905 \\
\textit{big-single} & 0.972 & 0.926 
& 0.917 \\
\textit{big-cv} & \textbf{0.976} & 0.933 
& \textbf{0.921} \\
\textit{big-cv-culled} & 0.972 & \textbf{0.951} 
& 0.915 \\
\hline
\textit{TF-IDF} & 0.978 & 0.971  
& 0.94 \\
\textit{best} & -- & 0.979 & 0.991 \\

\bottomrule
\end{tabular}
\end{table}

The final results are shown in Table \ref{tab:final}.
In general, one can observe $\mathrm{F_{0.5u}} > \mathrm{F_1} > \textrm{C@1}$ which is probably due to FN > FP and TP > TN and consequently a higher recall of MGTs.

\begin{table}[]
\caption{Detailed performance of \textit{big-cv} model (top three rows) on test sets against the best baseline and the best contribution. (a) The main test set without obfuscation, (b) test set incorporating most of the ELOQUENT obfuscation contributions, (c) final evaluation (macro-averages over all individual datasets).}
\label{tab:final}
\begin{tabular}{lcccccccc}
\toprule
\textbf{Evaluation set} & \textbf{ROC-AUC} & \textbf{Brier} & \textbf{C@1} & $\mathbf{F_1}$
 & $\mathbf{F_{0.5u}}$ & \textbf{Mean} & \textbf{FPR} & \textbf{FNR} \\
\midrule
\textit{(a) Test} & 0.958 & 0.912 & 0.882 & 0.911 & 0.943 & - & - & - \\
\textit{(b) ELOQUENT-01} & 0.884 & 0.749 & 0.625 & 0.746 & 0.877 & - & - & - \\
\textit{(c) Final} & 0.904 & 0.866 & 0.846 & 0.891 & 0.933 & 0.897 & 0.124 & 0.150 \\
\hline
\textit{Final TF-IDF} & 0.963 & 0.900 & 0.897 & 0.904 & 0.946 & 0.922 & 0.106 & 0.093 \\
\textit{Final best} & 0.995 & 0.984 & 0.982 & 0.989 & 0.993 & 0.989 & 0.006 & 0.018 \\
\bottomrule
\end{tabular}
\end{table}


\section{Conclusion}
Two general observations are: (1) larger capacity of boosted trees increased the detection performance, and (2) obfuscation considerably reduced it,
Although the our model have not reached the baseline TF-IDF scores,
in the outlook, the boosted trees have the capacity to learn on a larger number of features, so incorporating TF-IDF features~\cite{lorenz_baselineavengers_2024} or standardising feature frequencies, found to be greatly effective in stylometry~\cite{burrows_delta_2002,eder2016RJStylometry}, and other classic feature engineering techniques could be beneficial.
The straightforward augmentation of the training set with obfuscated samples can further improve the results.
The other unexplored avenue is simply hyperparameter optimisation (both in terms of feature set and LGBM parameters).
The main computational overhead in our method is feature extraction on the large training dataset.
Classifier training (and training continuation), inference and explanation~\cite{lundberg2020local2global} is inexpensive.
In summary, we perceive it as a trade-off between the smaller cost and greater explainability of boosted trees and the better generalisation of neural-based systems.

\begin{acknowledgments}
The research for this publication has been supported by a grant from the Priority Research Area DigiWorld under the Strategic Programme Excellence Initiative at Jagiellonian University.
JKO's research on the stylometric pipeline 
was financed by European Funds for Smart Economy, FENG program, CLARIN – Common Language Resources
and Technology Infrastructure, project no.FENG.02.04-IP.040004/24-00.

MM and TB participated in the submission as a programming assignment from the “AI Workshop II” course at Jagiellonian University during the summer term of 2025.
\end{acknowledgments}


\section*{Declaration on Generative AI}
  
 During the preparation of this work, the authors used Writefull's model in order to: Grammar and spelling check.
 After using this tool, the authors reviewed and edited the content as needed and take full responsibility for the publication’s content. 

\bibliography{stylo.bib}



\end{document}